\documentclass{article}
\usepackage[final]{graphicx}
\usepackage{algorithm}
\usepackage{algorithmic}
\usepackage{amssymb}
\usepackage{amsmath}
\usepackage{booktabs}
\usepackage{tikz}
\usetikzlibrary{arrows.meta, positioning, fit, backgrounds, calc}
\usepackage[preprint]{corl_2025} 

\DeclareMathOperator*{\argmin}{arg\,min}
\DeclareMathOperator*{\argmax}{arg\,max}

\title{Temporally Consistent Unsupervised Segmentation for Mobile Robot Perception}

\author{
\textbf{Christian C. Ellis}$^{1, 2}$ \quad
\textbf{Maggie B. Wigness}$^{2}$ \quad
\textbf{Craig T. Lennon}$^{2}$ \quad
\textbf{Lance Fiondella}$^{3}$ \\
$^{1}$Oden Institute for Computational Engineering \& Sciences, University of Texas at Austin \\
$^{2}$DEVCOM Army Research Laboratory, Adelphi, MD, United States \\
$^{3}$Department of Electrical and Computer Engineering, University of Massachusetts Dartmouth \\
\texttt{christian.ellis@austin.utexas.edu}, 
\texttt{maggie.b.wigness.civ@army.mil}, \\
\texttt{craig.t.lennon.civ@army.mil}, 
\texttt{lfiondella@umassd.edu}
}

\begin{document}
\maketitle

\begin{abstract}
\label{abstract}
Rapid progress in terrain-aware autonomous ground navigation has been driven by advances in supervised semantic segmentation. 
However, these methods rely on costly data collection and labor-intensive ground truth labeling to train deep models.
Furthermore, autonomous systems are increasingly deployed in unrehearsed, unstructured environments where no labeled data exists and semantic categories may be ambiguous or domain-specific.
Recent zero-shot approaches to unsupervised segmentation have shown promise in such settings but typically operate on individual frames, lacking temporal consistency—a critical property for robust perception in unstructured environments.
To address this gap we introduce Frontier-Seg, a method for temporally consistent unsupervised segmentation of terrain from mobile robot video streams. 
Frontier-Seg clusters superpixel-level features extracted from foundation model backbones—specifically DINOv2—and enforces temporal consistency across frames to identify persistent terrain boundaries or frontiers without human supervision.
We evaluate Frontier-Seg on a diverse set of benchmark datasets—including RUGD and RELLIS-3D-demonstrating its ability to perform unsupervised segmentation across unstructured off-road environments.
\end{abstract}


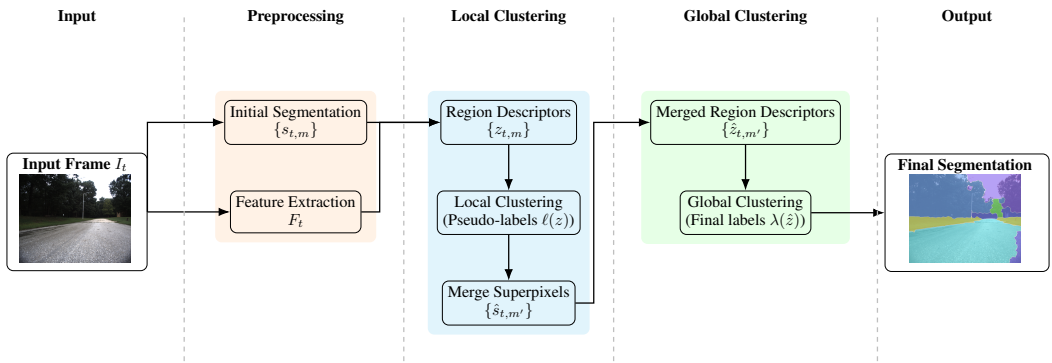
\begin{figure}[h]
\centering
\resizebox{\linewidth}{!}{
\begin{tikzpicture}[
    node distance=1.0cm and 1.6cm,
    box/.style = {draw, rounded corners, minimum width=2.6cm, minimum height=0.8cm, align=center, font=\normalsize},
    arrow/.style = {draw, -{Latex[length=2.5mm]}, thick}
    ]

\node[box] (features) {Feature Extraction\\$F_t$};
\node[box, above=of features] (segmentation) {Initial Segmentation\\$\{s_{t,m}\}$};
\node[box, left=1.8cm of features, inner sep=2pt] (input) {
  \begin{tabular}{c}
    \textbf{Input Frame $I_t$} \\
    \includegraphics[width=2.4cm]{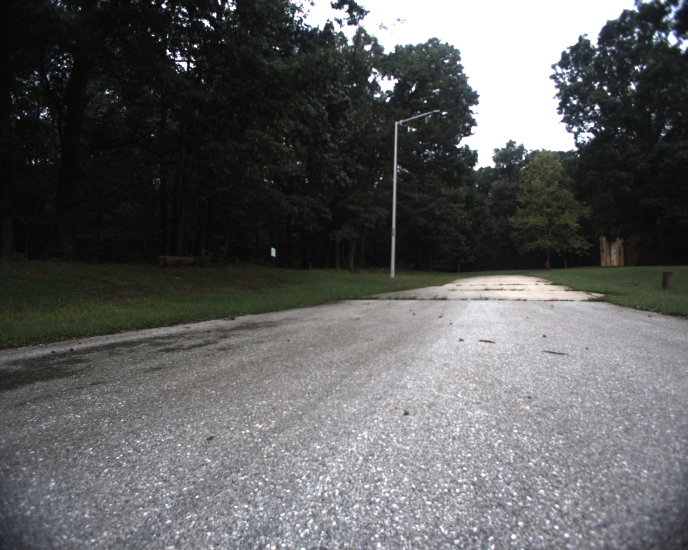}
  \end{tabular}
};

\node[box, right=1.6cm of segmentation] (descriptors) {Region Descriptors\\$\{z_{t,m}\}$};
\node[box, below=of descriptors] (local) {Local Clustering\\(Pseudo-labels $\ell(z)$)};
\node[box, below=of local] (merge) {Merge Superpixels\\$\{\hat{s}_{t,m'}\}$};

\node[box, right=1.6cm of descriptors] (merged_desc) {Merged Region Descriptors\\$\{\hat{z}_{t,m'}\}$};
\node[box, below=of merged_desc] (global) {Global Clustering\\(Final labels $\lambda(\hat{z})$)};

\node[box, right=1.6cm of global, inner sep=2pt] (output) {
  \begin{tabular}{c}
    \textbf{Final Segmentation} \\
    \includegraphics[width=2.4cm]{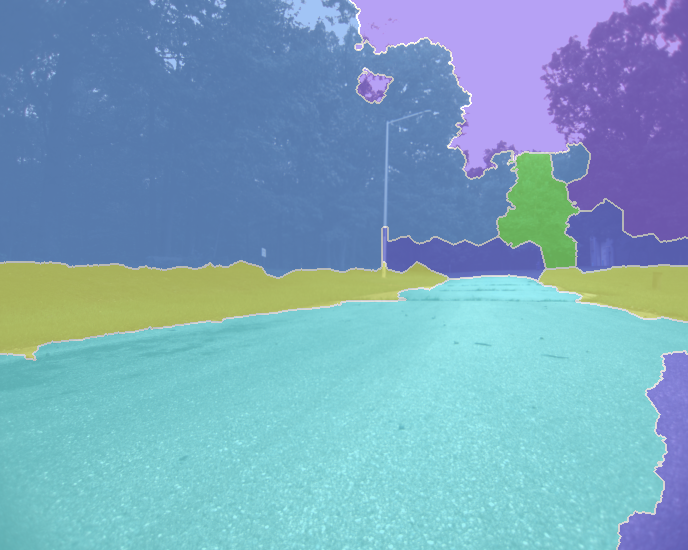}
  \end{tabular}
};
\draw[arrow] (input.east) -- (features.west);
\draw[arrow] (input.east) |- (segmentation.west);

\draw[arrow] (segmentation.east) -- (descriptors.west);
\draw[arrow] (features.east) -- ++(0.4,0) |- (descriptors.west);
\draw[arrow] (descriptors) -- (local);
\draw[arrow] (local) -- (merge);
\draw[arrow] (merge.east) -- ++(0.4,0) |- (merged_desc.west);
\draw[arrow] (merged_desc) -- (global);
\draw[arrow] (global) -- (output);

\begin{pgfonlayer}{background}
    \node[fit=(segmentation)(features), fill=orange!10, rounded corners, inner sep=0.2cm] (preprocessinggroup) {};
    \node[fit=(descriptors)(local)(merge), fill=cyan!10, rounded corners, inner sep=0.2cm] (localgroup) {};
    \node[fit=(merged_desc)(global), fill=green!10, rounded corners, inner sep=0.2cm] (globalgroup) {};
\end{pgfonlayer}

\coordinate (dashedtop) at ([yshift=1.8cm]segmentation.north);
\coordinate (dashedbottom) at ([yshift=-0.8cm]merge.south);

\draw[dashed, thick, gray!50]
    ([xshift=0.8cm]input.east |- dashedtop) -- ([xshift=0.8cm]input.east |- dashedbottom);
\draw[dashed, thick, gray!50]
    ([xshift=0.8cm]segmentation.east |- dashedtop) -- ([xshift=0.8cm]segmentation.east |- dashedbottom);
\draw[dashed, thick, gray!50]
    ([xshift=0.8cm]descriptors.east |- dashedtop) -- ([xshift=0.8cm]descriptors.east |- dashedbottom);
\draw[dashed, thick, gray!50]
    ([xshift=0.8cm]merged_desc.east |- dashedtop) -- ([xshift=0.8cm]merged_desc.east |- dashedbottom);

\coordinate (titlebaseline) at ([yshift=1.8cm]segmentation.north);

\node at ([xshift=0cm]input.center |- titlebaseline) {\textbf{Input}};
\node at ([xshift=0cm]segmentation.center |- titlebaseline) {\textbf{Preprocessing}};
\node at ([xshift=0cm]descriptors.center |- titlebaseline) {\textbf{Local Clustering}};
\node at ([xshift=0cm]merged_desc.center |- titlebaseline) {\textbf{Global Clustering}};
\node at ([xshift=0cm]output.center |- titlebaseline) {\textbf{Output}};

\end{tikzpicture}
}
\caption{Overview of the Frontier-Seg pipeline. Local clustering aggregates region descriptors over short video windows to assign pseudo-labels, while global clustering merges and re-clusters these descriptors across the full sequence to obtain the final segmentation.}
\label{fig:frontier_seg_pipeline}
\end{figure}

\section{Introduction}
\label{sec:introduction}
Autonomous ground robots are increasingly being deployed in complex, off-road environments where terrain is irregular, unstructured, and unfamiliar.
Reliable terrain understanding is essential for navigation in such settings, yet conventional perception systems often depend on supervised semantic segmentation models trained on extensive, manually annotated datasets.
This reliance poses a fundamental limitation: supervised approaches do not scale well to novel domains where labeled data is scarce, and semantic boundaries may be ambiguous or context-specific.
Moreover, defining a universally valid set of semantic categories for unstructured environments is itself ill-posed—concepts such as ``trail," ``grass," or ``mud" can vary dramatically in appearance, meaning, and navigational relevance depending on environmental conditions and operational context. Additionally, some terrain is likely better represented as a set of mixed semantics, e.g., forest terrain is composed of grass, dirt, twigs, and leaves.
Thus, to operate effectively in these scenarios, autonomous systems must adopt perception strategies that generalize beyond fixed taxonomies and can adapt to new environments quickly.

While supervised semantic segmentation~\cite{garcia2018survey, lateef2019survey} has enabled fine-grained scene understanding in structured environments such as urban navigation—where lane markings, pedestrians, and traffic signs define clear semantic boundaries~\cite{van2018autonomous}—these methods face critical limitations in off-road or semi-structured domains.
This problem is particularly acute in application domains such as humanitarian assistance and disaster relief~\cite{murphy2014disaster, nagatani2013emergency}, agriculture~\cite{reina2017terrain}, and forestry~\cite{la2024exploring} where terrain can be irregular and ambiguous. 
Off-road autonomous driving datasets~\cite{szabo2024comprehensive} have improved coverage of terrain variability, yet still focus ground truth perception annotations to support supervised approaches~\cite{feng2020deep, guo2019safe}, which hinge on large, labeled datasets with inherently constrained and fixed ontologies. 
Recent unsupervised and zero-shot segmentation methods~\cite{couairon2024diffcut, wang2023cut, shuai2024diffseg} offer an alternative by leveraging foundation model features to segment scenes without annotations.
However, these methods typically operate on single frames, ignoring the temporal continuity of robot video streams and resulting in fragmented, inconsistent segmentations over time—an issue that compromises downstream navigation and planning reliability~\cite{varghese2020unsupervised}.

To address these challenges, we introduce Frontier-Seg, a method for temporally consistent unsupervised segmentation from mobile robot video streams.
Frontier-Seg centers on a two-phase clustering process. First, region descriptors are extracted and clustered for a set of different temporal windows to assign initial pseudo-labels locally within the data. Second, local psuedo-labels are used to recompute and refine region descriptors, which are then aggregated across the local windows to globally cluster data to produce temporally consistent psuedo-labels across the video stream.
This design enforces temporal consistency without requiring motion cues or explicit tracking, allows adaptive region discovery without a fixed semantic ontology, and produces coherent segmentations that persist across robot motion through unstructured environments.


Our main contributions are as follows.
(1) We propose Frontier-Seg, a method that clusters temporally aggregated superpixel-level features extracted from foundation models to discover persistent terrain structures, enabling users to define the granularity of the ontology without prescribing specific semantic categories.
(2) We introduce a temporal windowing and feature aggregation strategy that enforces region consistency across frames without requiring explicit motion estimation or supervision.
(3) We present a region descriptor recomputation mechanism that refines segmentation quality by aligning features with evolving spatial structure.
(4) We demonstrate through extensive experiments on the RUGD~\cite{wigness2019rugd} and RELLIS-3D~\cite{jiang2021rellis} datasets that Frontier-Seg achieves strong unsupervised segmentation performance in challenging, off-road environments, establishing a new foundation for robust terrain perception without human labels.

By eliminating the need for manual labels and rigid taxonomies, and by enhancing temporal stability, Frontier-Seg provides a scalable foundation for terrain understanding in domains where rapid deployment and adaptability are critical.
As such, Frontier-Seg represents an important step toward autonomous systems that can continuously adapt their perception models to novel, unstructured settings without external supervision.

\section{Related Work}
\label{sec:related_work}

\paragraph{Supervised Semantic Segmentation.}
Semantic segmentation produces dense pixel-wise labeling that provides environmental context about terrain and objects in the scene that can be used for high level reasoning and planning.
The advances in this space using deep learning architectures~\cite{garcia2018survey, lateef2019survey} has carried over into the autonomous vehicle domain~\cite{van2018autonomous} where processing of perception semantics are used to support traversability analysis~\cite{meng2023terrainnet}, robotic behavior learning~\cite{miyamoto2019vision, Wigness-2021-131288}, and uncertainty aware path planning~\cite{ellis2021risk, tan2021risk}.
The limitation of these supervised approaches is the need for large annotated datasets~\cite{feng2020deep, guo2019safe, liu2024survey, szabo2024comprehensive}, and the inability to generalize to open-world settings~\cite{parmar2023open} given they have learned a fixed ontology.
Yet, for the underlying motivating application of off-road navigation in unseen environments, the importance of open-world or zero-shot semantic segmentation is critical.

\paragraph{Unsupervised Semantic Segmentation.} Unsupervised segmentation emerged as a way to make image processing more efficient by moving from pixel-wise computations to segment-based computations. These early segmentation approaches relied on low-level cues such as color, texture, and edge information to define pixel affinities, and applied greedy or spectral partitioning to group pixels into perceptually coherent regions~\cite{shi2000normalized, comaniciu2002mean, felzenszwalb2004efficient}.
Concurrent frameworks introduced energy-based models that optimized global objectives to enforce region homogeneity~\cite{chan2001active} and boundary alignment~\cite{boykov2004experimental}.
Hierarchical approaches improved performance through multiscale boundary detection and region merging~\cite{arbelaez2010contour}, while fast oversegmentation techniques focused on generating compact, spatially regular superpixels~\cite{achanta2012slic}. Output of these approaches were largely still over-segmented  with respect to ground truth semantic concepts with options for hierarchical output to meet the varying segment granularity needs for different downstream tasks.

With the rise of deep learning, unsupervised segmentation techniques were advanced to more specifically focus on segmenting with respect to ground truth semantics through methods that combined self-supervised feature learning with clustering~\cite{caron2018deep, ji2019invariant}, part discovery~\cite{hung2019scops}, and equivariance constraints~\cite{cho2021picie}.
Most similar to our work, the latest approaches leverage pretrained vision transformers~\cite{oquab2024dinov, zhou2021ibot, he2022masked} for dense affinity modeling~\cite{hamilton2022unsupervised}, as well as diffusion-based mechanisms to propagate semantic signals across spatial regions~\cite{shuai2024diffseg, couairon2024diffcut}. However, these approaches typically operate on individual frames to support zero-shot semantic segmentation.

\paragraph{Temporal Consistency in Video Segmentation.}
Enforcing coherence across video frames is a longstanding challenge. Use of optical flow networks~\cite{weinzaepfel2013deepflow, revaud2015epicflow, ilg2017flownet} have been used to support extension of image-based semantic segmentation to video sequences~\cite{zhu2017deep}, but its reliance on accurate motion estimation limits robustness especially in environments with high occlusion.
To address inconsistent optical flow while ensuring temporal semantic consistency, a motion state alignment network~\cite{su2023motion} and a temporal memory attention module~\cite{wang2021temporal}—which captures temporal feature correlations from image sequences without the overhead of explicit flow computation—were introduced.
Yet, these approaches still leverage supervised semantic segmentation networks, trained on finite ontologies, to ensure consistent semantics are propagated throughout the video sequence.

Unsupervised video segmentation approaches tend to focus on object-centric segmentation~\cite{gao2023deep}, which fails to provide dense labeling for terrain and other background concepts that are relevant for autonomous navigation.
Or, similar to the image domain, they lack semantic focus as the underlying objective is to provide pre-processing segment capabilities for video processing~\cite{xu2012streaming}. 
Similar to our work,an unsupervised segmentation framework for streaming data~\cite{wigness2017unsupervised} similarly used local and global clustering, but required multiple passes per frame to produce ensembled output.

\paragraph{Frontier-Seg in Context.}
In contrast to prior work, \textit{Frontier-Seg} performs unsupervised segmentation using a 2-phase clustering scheme (locally across a regional window of frames and globally across the video sequence) by leveraging DINOv2 features and SLIC-based superpixels, resulting in a post-clustering refinement that enforces temporal consistency without explicit tracking or motion cues.
By aligning descriptors to evolving spatial structure, it produces coherent terrain groupings well-suited for mobile robots navigating unstructured environments.

\clearpage
\section{Methodology}
\label{sec:methodology}
Frontier-Seg provides terrain-aware perception for autonomous navigation in an unsupervised manner by identifying a representation of visual concepts in the environment using stream-based unsupervised segmentation.

\subsection{Problem Formulation}
We address the problem of \textit{unsupervised, temporally consistent terrain segmentation} from video streams collected by mobile ground robots operating in \textit{unstructured outdoor environments}.  
Formally, given a sequence of RGB frames $\{I_t\}_{t=1}^T$, captured over time from a robot's onboard camera, the objective is to assign a pseudo-label $y_{p,t} \in \{1, \dots, K\}$ to each pixel $p$ in each frame $I_t$ \textit{without any human supervision} (i.e., no ground-truth annotations).
Additionally, the semantic assignments must exhibit \textit{temporal consistency}: regions representing the same terrain class should maintain coherent labels across adjacent frames despite appearance changes, motion, and viewpoint shifts.
The challenge is compounded by the lack of predefined ontologies, the high variability of unstructured terrains, and the potential ambiguity between similar textures or visual patterns.
Thus, the method must both \textit{discover} terrain classes in an unsupervised manner and \textit{track} them consistently over time, enabling robust perception without prior knowledge of the environment.

\subsection{Algorithm Overview}
Frontier-Seg addresses the problem of unsupervised, temporally consistent terrain segmentation through a three-stage pipeline: \textit{initial segmentation and feature extraction}, \textit{local clustering}, and \textit{global clustering}.
Given an input video stream, Frontier-Seg first applies two independent processes to each frame: initial segmentation to group spatially coherent regions based on low-level image information, and dense per-pixel feature extraction using a vision foundation model backbone.  
The initial segment mask and extracted features are then jointly used to compute compact feature descriptors for each region by pooling features within segment boundaries, which we call region descriptors.
Within each temporal window, local clustering is performed over the region descriptors to assign preliminary pseudo-labels.  
To enforce consistency over time, a global clustering stage merges local clusters across frames, aligning pseudo-labels based on feature similarity and temporal correspondence.
This two-stage clustering strategy enables the system to both discover terrain classes without supervision and maintain stable segmentation across video sequences.
Additional details of these stages are presented in the rest of this section.

\subsection{Initial Segmentation and Feature Extraction}
The first stage of Frontier-Seg applies initial segmentation and feature extraction independently to each input frame $I_t \in \mathbb{R}^{H \times W \times 3}$.  
Initial segmentation partitions the image into a set of $M_t$ non-overlapping segments, $\{s_{t,m}\}_{m=1}^{M_t}$, where each $s_{t,m} \subseteq \{1, \dots, H\} \times \{1, \dots, W\}$ denotes the set of pixel indices belonging to the $m$-th segment at time $t$.  
We employ SLIC superpixels~\cite{achanta2012slic} to generate these segments based on low-level image cues such as color and spatial proximity.

In parallel, we extract dense per-pixel features $F_t \in \mathbb{R}^{H \times W \times D}$ using a vision foundation model backbone, where $D$ denotes the feature dimensionality (final layer of the foundation model).
Each feature vector $f_{t,p} \in \mathbb{R}^{D}$ corresponds to pixel $p$ in frame $I_t$.
Specifically, we employ the DINOv2~\cite{oquab2024dinov} vision foundation model as the feature backbone, leveraging its ability to produce semantically rich and spatially consistent feature maps without requiring supervision.  

Given the initial segment mask and the feature map, we compute a compact region descriptor $z_{t,m} \in \mathbb{R}^{D}$ for each segment $s_{t,m}$ by masked average pooling:
\begin{equation}
z_{t,m} = \frac{1}{|s_{t,m}|} \sum_{p \in s_{t,m}} f_{t,p}
\label{eq:initial_pooling}
\end{equation}
where $|s_{t,m}|$ denotes the number of pixels in segment $s_{t,m}$.
The set of region descriptors $\{z_{t,m}\}_{m=1}^{M_t}$ for each frame forms the input to the subsequent local clustering stage.


\subsection{Local Clustering}
Given the set of region descriptors $\{z_{t,m}\}_{m=1}^{M_t}$ extracted from each frame $I_t$, the goal of the local clustering stage is to assign a preliminary pseudo-label to each segment based on its feature representation.  
To do so, we aggregate region descriptors across a temporal window of frames and perform clustering in feature space.

Specifically, let $\mathcal{Z}_w = \{ z_{t,m} \}$ denote the collection of all region descriptors extracted from frames within a local window $w = \{t, t+1, \dots, t+\Delta t\}$.  
We apply $K$-means clustering to $\mathcal{Z}_w$, partitioning the descriptors into $K$ clusters based on Euclidean distance in feature space.  
The cluster assignment function is defined as:
\begin{equation}
\ell(z) = \argmin_{k \in \{1, \dots, K\}} \| z - \mu_k \|_2^2
\label{eq:local_clustering}
\end{equation}
where $\mu_k$ denotes the centroid of the $k$-th cluster.

Each region descriptor $z_{t,m}$ is thus assigned a preliminary pseudo-label $\ell(z_{t,m}) \in \{1, \dots, K\}$, producing an initial segmentation of the scene within the local temporal window by grouping regions with similar semantic and structural characteristics.  
Based on these assignments, we merge the initial superpixels within each frame according to their pseudo-labels, resulting in a new set of merged segments $\{ \hat{s}_{t,m'} \}_{m'=1}^{\hat{M}_t}$, where $\hat{M}_t \leq M_t$.

For each merged segment $\hat{s}_{t,m'}$, we recompute a new region descriptor $\hat{z}_{t,m'} \in \mathbb{R}^{D}$ by masked average pooling over the dense feature map:
\begin{equation}
\hat{z}_{t,m'} = \frac{1}{|\hat{s}_{t,m'}|} \sum_{p \in \hat{s}_{t,m'}} f_{t,p}
\label{eq:merged_pooling}
\end{equation}

These recomputed region descriptors $\{ \hat{z}_{t,m'} \}_{m'=1}^{\hat{M}_t}$ form the input to the subsequent global clustering stage.  
The local clustering stage operates independently across non-overlapping temporal windows, producing merged region descriptors that serve as the input for global clustering, which enforces label consistency across the full video sequence.

\subsection{Global Labeling}
While local clustering provides preliminary pseudo-labels within a temporal window, it does not guarantee consistency of labels across windows.  
To enforce temporal consistency across the full video sequence, we introduce a global clustering stage that merges the merged regions into globally consistent terrain classes.

We collect all recomputed region descriptors $\{ \hat{z}_{t,m'} \}$ from every frame in the sequence.  
We then apply $K$-means clustering globally over this aggregated set, partitioning the merged region descriptors into $K$ globally defined clusters.  
The global cluster assignment function is defined as:
\begin{equation}
\lambda(\hat{z}) = \argmin_{k \in \{1, \dots, K\}} \| \hat{z} - \nu_k \|_2^2
\label{eq:global_clustering}
\end{equation}
where $\nu_k$ denotes the centroid of the $k$-th global cluster.

Each merged region $\hat{s}_{t,m'}$ is thus assigned a final global label $\lambda(\hat{z}_{t,m'}) \in \{1, \dots, K\}$.  
The final per-pixel segmentation is obtained by propagating the global label of each merged region to all pixels it contains.  
This global clustering stage completes the segmentation process, enabling temporally stable, unsupervised terrain understanding across the video sequence.

\clearpage


\section{Experiments}
\label{sec:experiments}
We evaluate Frontier-Seg on the RUGD~\cite{wigness2019rugd} and RELLIS-3D~\cite{jiang2021rellis} datasets to quantify performance and demonstrate applicability to temporally consistent unsupervised segmentation of video streams for mobile robot perception.
RUGD contains $18$ temporal sequences ($49$ to $849$ frames)
,while RELLIS-3D contains $5$ sequences ($900$ to $2074$ frames).
Although designed for flexible semantic output, Frontier-Seg is assessed under a standard supervised segmentation framework with metrics including mIoU, pixel accuracy, and over- and under-segmentation entropy.
Ground truth annotations are used solely for evaluation and are never seen during training.

\subsection{Experimental Setup}
Frontier-Seg is implemented using the \texttt{facebook/dinov2-with-registers-base}~\cite{darcet2023vision} Vision Transformer as the feature backbone.
Each RGB image is resized to $512$×$512$ and processed to extract dense per-pixel feature vectors of dimension $768$.
Prior to segmentation, each image is converted to the CIELAB color space and smoothed with a Gaussian blur ($\sigma = 0.7$) to reduce noise and improve superpixel coherence. 
Initial segmentation is then performed using SLICO~\cite{achanta2012slic}, with a region size of 30, resulting in approximately 200 superpixels per frame.
For temporal modeling, videos are divided into non-overlapping windows of 100 consecutive frames ($\Delta t = 99$). 
Within each window, a region descriptor is computed for each superpixel by (1) performing masked average pooling over dense per-pixel features within the superpixel mask, (2) aggregating register tokens via attention conditioned on the pooled feature, and (3) blending the result with the CLS token to form the final region descriptor.
Local clustering is performed via $K$-means with $K = 100$ to assign pseudo-labels within each window. After region merging based on label agreement, updated descriptors are recomputed.
For global clustering, all descriptors across all temporal windows are aggregated and clustered again using $K$-means with $K = 50$ to assign globally consistent labels.

We re-evaluate DiffCut~\cite{couairon2024diffcut} using the authors’ open-source implementation.
DiffCut uses the Segmind Stable Diffusion-1B (SSD-1B) model~\cite{gupta2024progressive} as a backbone and hyperparameter $\tau$ to adjust over-segmentation.
To achieve a similar level of over-segmentation as Frontier-Seg\textemdash we run DiffCut with two hyperparameter values, $\tau=0.9, 0.95$. 
For a fair comparison, we also present results for a version of Frontier-Seg utilizing the same SSD-1B backbone.

We report quantitative results under two evaluation settings: \textbf{zero-shot} (frame-by-frame) and \textbf{temporal}.
In the zero-shot setting, segmentations are evaluated independently per frame without temporal context.
DiffCut~\cite{couairon2024diffcut} is evaluated using per-frame predictions, which are aligned to the ground truth via many-to-one Hungarian matching based on region overlap.
For zero-shot evaluation of Frontier-Seg, we run the full temporal model but apply Hungarian matching on each frame independently to ensure comparability.
In the temporal setting, we assess the consistency and persistence of segmentations over time.
We adapt DiffCut by using its zero-shot segmentation output and SSD-1B features as input to our local and global clustering pipeline, enabling a fair comparison under a shared temporal modeling framework.
Unlike the zero-shot setup, we perform a single Hungarian matching over the full video sequence after temporal aggregation to evaluate long-term coherence.



\subsection{Quantitative Metrics}
Unsupervised segmentation performance is evaluated using two standard semantic segmentation metrics: mean Intersection-over-Union (mIoU) and mean pixel accuracy (Acc).
Because unsupervised segments rarely exhibit a one-to-one correspondence with ground truth labels, we follow prior work~\cite{couairon2024diffcut} and apply many-to-one Hungarian matching~\cite{kuhn1955hungarian} based on maximal overlap between predicted and ground truth regions before computing these metrics.
However, greater over-segmentation can artificially inflate mIoU and Acc under this matching scheme.
To address this, we additionally report over-segmentation entropy (OSE) and under-segmentation entropy (USE)~\cite{gong2011conditional}, which quantify the fragmentation and mixing between predicted and ground truth regions.
These metrics capture the trade-off between segmentation granularity and semantic compactness, while providing a more holistic view of segmentation quality in the unsupervised setting.

\clearpage

\subsection{Quantitative Results}
\begin{table*}[t]
\centering
\caption{Quantitative comparison on RUGD and RELLIS-3D, averaged across all subdatasets.}
\label{tab:quantitative_results}
\renewcommand{\arraystretch}{1.2}
\setlength{\tabcolsep}{3pt}
\begin{tabular}{l*{8}{r}}
\toprule
\textbf{Method} & \multicolumn{4}{c}{\textbf{Zero Shot}} & \multicolumn{4}{c}{\textbf{Temporal}} \\
\cmidrule(lr){2-5} \cmidrule(lr){6-9}
& \textbf{mIoU $\uparrow$} & \textbf{Acc $\uparrow$} & \textbf{OSE $\downarrow$} & \textbf{USE $\downarrow$} & \textbf{mIoU $\uparrow$} & \textbf{Acc $\uparrow$} & \textbf{OSE $\downarrow$} & \textbf{USE $\downarrow$} \\
\midrule
\textbf{RUGD} & & & & & & & & \\
DiffCut ($\tau=0.9$)~\cite{couairon2024diffcut} & 51.02 & 90.41 & 2.49 & 0.23 & 13.94 & 64.40 & 1.14 & 0.81\\
DiffCut ($\tau=0.95$)~\cite{couairon2024diffcut} & 53.93 & \textbf{91.70} & 3.53 & \textbf{0.19} & 13.29 & 59.15 & 1.19 & 0.84 \\
Frontier-Seg (DinoV2-B, Ours) & \textbf{56.00} & 89.60 & 2.42 & 0.26 & \textbf{34.15} & \textbf{81.76} & \textbf{0.98} & \textbf{0.39} \\
Frontier-Seg (SSD-1B, Ours) & 39.89 & 77.79 & \textbf{1.29} & 0.54 & 24.80 & 69.91 & 1.30 & 0.54 \\
\addlinespace
\textbf{RELLIS-3D} & & & & & & & & \\
DiffCut ($\tau=0.9$)~\cite{couairon2024diffcut} & 46.18 & 86.35 & 1.87 & 0.35 & 13.79 & 64.13 & \textbf{0.73} & 0.87 \\
DiffCut ($\tau=0.95$)~\cite{couairon2024diffcut} & \textbf{49.38} & \textbf{88.02} & 2.81 & \textbf{0.29} & 11.45 & 60.45 & 0.81 & 0.98 \\
Frontier-Seg (DinoV2-B, Ours) & 38.59 & 82.03 & 1.24 & 0.49 & \textbf{31.12}  & \textbf{80.82} & 1.25 & \textbf{0.45} \\
Frontier-Seg (SSD-1B, Ours) & 30.02 & 74.27 & \textbf{1.00} & 0.66  & 18.88 & 72.04 & 1.01 & 0.66 \\
\bottomrule
\end{tabular}
\end{table*}

Quantitative results are summarized in Table~\ref{tab:quantitative_results}. 
While DiffCut achieves higher zero-shot accuracy across both datasets (e.g., $91.70$ vs. $89.60$ on RUGD) and higher mIoU on RELLIS-3D ($49.38$ vs. $38.59$), Frontier-Seg is substantially more consistent temporally. 
Frontier-Seg demonstrates strong performance in temporally consistent unsupervised segmentation, achieving the highest temporal mIoU and accuracy on both RUGD ($34.15$, $81.76$) and RELLIS-3D ($31.12$, $80.82$), compared to DiffCut ($\tau=0.90$), which reaches ($13.94$, $64.40$) and ($13.79$, $64.13$), respectively. 
These results suggest that clustering temporally aggregated superpixel-level features is effective at capturing terrain structure in off-road video. 
Moreover, the consistency of Frontier-Seg’s performance across both datasets suggests that its temporal modeling approach generalizes well across diverse, unstructured terrain types.

In terms of temporal consistency, Frontier-Seg generally obtains lower over- and under-segmentation entropy, suggesting improved stability and coherence of segment boundaries over time. 
For example, on RUGD, OSE and USE are ($0.98$, $0.39$) for Frontier-Seg (DINOv2), compared to (1.14, 0.81) for DiffCut.
While DiffCut achieves a slightly lower OSE of $0.73$ on RELLIS-3D in one setting ($\tau=0.9$), its corresponding USE remains high ($0.87$), indicating limited temporal coherence overall.
Finally, the relatively high USE values observed in DiffCut ($0.84$ on RUGD and $0.98$ on RELLIS-3D) point to frequent fragmentation of semantic regions; high USE occurs when predicted segments span multiple ground truth classes, resulting in noisy pseudo-labels that undermine stable clustering.

Within Frontier-Seg, the ViT-based DINOv2 backbone consistently outperforms the SSD-1B variant on both RUGD and RELLIS-3D. 
For example, on RUGD, DINOv2 achieves higher zero-shot mIoU ($56.00$ vs. $39.89$) and temporal mIoU ($34.15$ vs. $24.80$),  suggesting that transformer-based features may offer advantages for modeling temporal structure in unsupervised terrain segmentation.
This performance gap highlights the importance of backbone selection when designing segmentation pipelines for off-road, temporally structured data.

In addition to improved segmentation quality, DINOv2 offered faster inference and easier integration into clustering pipelines.
DiffCut required 50 denoising steps per image (e.g., $\sim$1.2s for $512 \times 512$ on an RTX 4090), DINOv2 extracts dense features in a single pass ($\sim$250ms), making it more practical for real-time applications.
Together, these results support the effectiveness of Frontier-Seg’s design: combining superpixel-level aggregation, temporal feature pooling, and region refinement yields stable, scalable segmentation without supervision.
By leveraging DINOv2’s dense, purely visual features without dependence on generative diffusion priors or language conditioning, Frontier-Seg remains lightweight, broadly applicable, and well-suited for real-world deployment on resource-constrained mobile robotic platforms.

\clearpage
\subsection{Qualitative Results}
\begin{figure}
    \centering
    \includegraphics[width=0.85\textwidth]{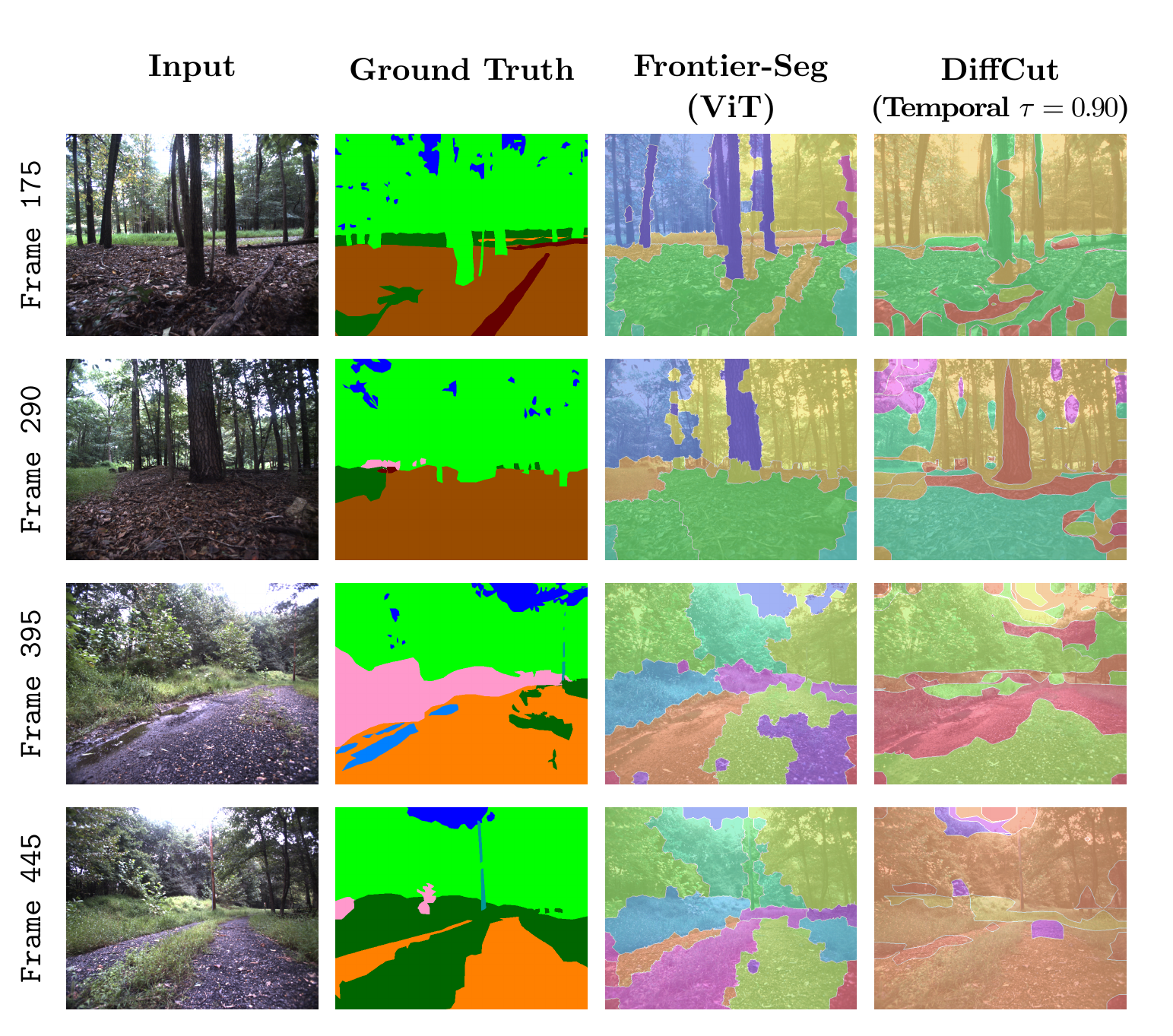}
    \caption{Qualitative comparison on four RUGD sequences. Columns show the input frame, ground truth annotation, and predictions from Frontier-Seg (ViT-based) and DiffCut (temporal). Our method demonstrates improved spatial alignment and semantic consistency under challenging terrain variations.}
    \label{fig:qualitative_grid}
\end{figure}

Qualitative results are found in Fig.~\ref{fig:qualitative_grid}.
Across all models, Frontier-Seg (ViT) produces the most spatially aligned and semantically coherent segments across all frames, while matching the ground truth boundaries the best.
Its delineation of fine-grained structures—such as tree trunks, path edges, and vegetation boundaries—is consistently sharper and less fragmented than the baseline.
Compared to DiffCut, which frequently merges disparate regions and exhibits temporal inconsistency, Frontier-Seg maintains stable, high-purity segments across challenging terrain transitions.
This coherence underscores Frontier-Seg’s capacity to capture structural detail while preserving consistency over time in complex off-road environments.




\section{Conclusion}
\label{sec:conclusion}
In this work, we introduced Frontier-Seg, a new method for temporally consistent unsupervised segmentation of mobile robot video streams.
By clustering dense foundation model features over superpixels and leveraging a novel temporal windowing and feature aggregation strategy, Frontier-Seg identifies persistent terrain boundaries without requiring motion estimation or human supervision.
Our region descriptor recomputation mechanism further refines spatial alignment across frames, improving segmentation quality.
Extensive experiments on the RUGD and RELLIS-3D datasets demonstrate that Frontier-Seg achieves strong unsupervised segmentation performance in challenging, off-road environments, setting a new foundation for robust, label-free terrain perception.
Future work will explore deploying Frontier-Seg online onto real-world robotic hardware and utilizing its perception output for autonomous navigation.
\clearpage

\section{Limitations}
Frontier-Seg operates in an offline setting and is not currently designed for streaming or online deployment.
The global clustering stage requires access to all region descriptors across a video sequence, necessitating complete observation of the sequence prior to global label assignment.
This limits applicability in time-sensitive or evolving environments where real-time adaptability is essential.
Future work could explore incremental or online clustering methods that preserve temporal consistency while supporting streaming input.

Additionally, the storage and clustering of region descriptors during global clustering remains the most computationally demanding stage, limiting deployment in resource-constrained settings.
Optimizing this stage for real-time performance with minimal degradation in segmentation quality would enable broader deployment on edge devices and support closed-loop autonomy in field robotics applications.

Frontier-Seg assumes smooth frame-to-frame continuity and consistent egomotion. 
These assumptions may break down in highly dynamic scenes or under rapid viewpoint changes, potentially leading to segmentation drift or degraded performance. 
Improving robustness to abrupt motion or disordered temporal input remains an open direction.


\acknowledgments{
Research reported in this paper was sponsored in part by the DEVCOM Army Research Laboratory under Cooperative Agreement W911NF23-2-0211.
The views and conclusions contained in this document are those of the authors and should not be interpreted as representing the official policies, either expressed or implied, of the DEVCOM Army Research Laboratory or the U.S. Government.
The U.S. Government is authorized to reproduce and distribute reprints for Government purposes notwithstanding any copyright notation herein.
}

\bibliography{references}
\clearpage

\appendix
\begin{center}
    {\LARGE\bfseries
    Temporally Consistent Unsupervised Segmentation for Mobile Robot Perception \\
    \vspace{1em}
    Supplementary Material \par}
    \vspace{2em}
    {
    \textbf{Christian C. Ellis}$^{1, 2}$ \quad
    \textbf{Maggie B. Wigness}$^{2}$ \quad
    \textbf{Craig T. Lennon}$^{2}$ \quad
    \textbf{Lance Fiondella}$^{3}$ \\
    $^{1}$Oden Institute for Computational Engineering \& Sciences, University of Texas at Austin \\
    $^{2}$DEVCOM Army Research Laboratory, Adelphi, MD, United States \\
    $^{3}$Department of Electrical and Computer Engineering, University of Massachusetts Dartmouth \\
    \texttt{christian.ellis@austin.utexas.edu}, 
    \texttt{maggie.b.wigness@army.mil}, \\
    \texttt{craig.t.lennon@army.mil}, 
    \texttt{lfiondella@umassd.edu}
    \par
    }
\end{center}

\section{Overview of Supplementary Material}
This supplementary document provides additional technical details, evaluation methodology, and extended quantitative results to support the main paper, Temporally Consistent Unsupervised Segmentation for Mobile Robot Perception. 
Section 2 elaborates on the Frontier-Seg algorithm with high-level pseudocode and implementation details. 
Section 3 presents our evaluation strategy, including label alignment, performance metrics, and extensive quantitative results across both the RUGD and RELLIS-3D datasets under temporal and zero-shot settings. 
We further analyze performance trends across varying numbers of clusters and provide detailed over- and under-segmentation entropy metrics to better capture structural segmentation quality.

The supplementary results go beyond what is feasible to show in the main paper, providing detailed empirical evidence for design decisions such as the choice of backbone, and the impact of clustering granularity on both semantic accuracy and structural coherence.
We report mean Intersection over Union (mIoU), pixel accuracy (Acc), and entropy-based measures of over- and under-segmentation (OSE/USE), which jointly characterize both semantic accuracy and structural coherence.
The included tables compare Frontier-Seg with DiffCut~\cite{couairon2024diffcut}, across both temporal and zero-shot segmentation regimes.
DiffCut is included as a strong baseline due to its recent success in zero-shot semantic segmentation and its conceptual similarity to our region-based approach.
This expanded evaluation reveals systematic trends in performance variation and helps diagnose algorithmic behavior under different deployment settings.

Subsequent tables are organized by dataset, evaluation mode (temporal vs. zero-shot), and metric. 
Readers are encouraged to examine trends across rows (subdatasets) and columns (cluster count) to assess label stability, segmentation fidelity, and the effect of feature backbone and windowing strategy.
Notably, varying the number of clusters $K$ controls the granularity of the global label ontology: larger $K$ values yield finer partitions that may better capture small or rare structures but increase fragmentation risk, as reflected in higher over-segmentation entropy.
Conversely, smaller $K$ values promote compact representations but may overlap semantically distinct regions.
These trade-offs are visible in both performance metrics and entropy scores, underscoring the importance of selecting $K$ to balance discriminative power with structural coherence.

Note that for the zero-shot DiffCut models, we do not vary the number of clusters $K$.
This is because DiffCut automatically determines the number of segments—and thus the size of the ontology—on a per-frame basis.
For the temporal setting, we adapt DiffCut by leveraging its zero-shot segmentation outputs as region proposals and using SSD-1B features as the embedding space.
These region descriptors are then passed through our local and global clustering stages, enabling temporal consistency across frames.
While the number of initial segments varies per frame due to DiffCut’s frame-wise operation, the downstream clustering stages operate over a fixed number of clusters $K$, thereby standardizing the global label ontology across the video.
\newpage

\section{Methodology}
\subsection{Algorithm Overview}
To complement the methodology described in Section 3, we provide a high-level pseudocode summary of the Frontier-Seg algorithm.
The method operates on a sequence of video frames and consists of three main stages:
(1) initial superpixel segmentation and region-level feature extraction using dense descriptors,
(2) local clustering within temporal windows followed by intra-frame merging to refine regions,
and (3) global clustering of region descriptors to enforce consistent pseudo-labels across time.
The result is a temporally coherent, unsupervised segmentation of terrain that evolves smoothly across video frames.

\begin{algorithm}[H]
\caption{Frontier-Seg algorithm for temporally consistent unsupervised terrain segmentation. Region descriptors are computed, clustered locally, merged, and finally clustered globally to produce consistent segmentation across video frames.}
\label{alg:frontier_seg}
\begin{algorithmic}[1]
\REQUIRE Video frames $\{I_t\}_{t=1}^T$
\ENSURE Per-pixel segmentation maps with temporally consistent labels

\textbf{Step 1: Initial Segmentation and Feature Extraction}
\FOR{each frame $I_t$}
    \STATE Perform initial segmentation to obtain superpixels $\{s_{t,m}\}_{m=1}^{M_t}$
    \STATE Extract dense per-pixel features $F_t \in \mathbb{R}^{H \times W \times D}$
    \FOR{each segment $s_{t,m}$}
        \STATE Compute region descriptor $z_{t,m}$ using masked average pooling (Eq.~\eqref{eq:initial_pooling_repeat})
    \ENDFOR
\ENDFOR

\textbf{Step 2: Local Clustering and Region Merging}
\FOR{each temporal window $w = \{t, \dots, t+\Delta t\}$}
    \STATE Collect descriptors $\mathcal{Z}_w = \{ z_{t,m} \}$ within window $w$
    \STATE Perform $K$-means clustering to assign preliminary pseudo-labels $\ell(z)$ (Eq.~\eqref{eq:local_clustering_repeat})
    \STATE Merge superpixels within each frame according to $\ell(z_{t,m})$ to form merged segments $\{ \hat{s}_{t,m'} \}$
    \FOR{each merged segment $\hat{s}_{t,m'}$}
        \STATE Recompute merged region descriptor $\hat{z}_{t,m'}$ using masked average pooling (Eq.~\eqref{eq:merged_pooling_repeat})
    \ENDFOR
\ENDFOR

\textbf{Step 3: Global Clustering for Temporal Consistency}
\STATE Aggregate all merged region descriptors $\{ \hat{z}_{t,m'} \}$ across sequence
\STATE Perform global $K$-means clustering to assign final labels $\lambda(\hat{z})$ (Eq.~\eqref{eq:global_clustering_repeat})

\FOR{each frame $I_t$}
    \FOR{each pixel $p$}
        \STATE Assign pixel $p$ the global label of its corresponding merged segment
    \ENDFOR
\ENDFOR
\end{algorithmic}
\end{algorithm}

\begin{equation}
z_{t,m} = \frac{1}{|s_{t,m}|} \sum_{p \in s_{t,m}} f_{t,p}
\label{eq:initial_pooling_repeat}
\end{equation}

\begin{equation}
\ell(z) = \argmin_{k \in \{1, \dots, K\}} \| z - \mu_k \|_2^2
\label{eq:local_clustering_repeat}
\end{equation}

\begin{equation}
\hat{z}_{t,m'} = \frac{1}{|\hat{s}_{t,m'}|} \sum_{p \in \hat{s}_{t,m'}} f_{t,p}
\label{eq:merged_pooling_repeat}
\end{equation}

\begin{equation}
\lambda(\hat{z}) = \argmin_{k \in \{1, \dots, K\}} \| \hat{z} - \nu_k \|_2^2
\label{eq:global_clustering_repeat}
\end{equation}

\newpage

\section{Quantitative Results}

\subsection{Hungarian Matching for Cluster-to-Class Alignment}

To evaluate segmentation performance in the unsupervised setting, we must align abstract cluster labels with semantic ground-truth classes.

Let \( \mathcal{P} = \{p_1, \dots, p_N\} \) denote the set of all pixels in the evaluation set (either a single frame or an entire video sequence).  
Let \( y(p_i) \in \{1, \dots, C\} \) be the ground-truth class label for pixel \( p_i \), and let \( \hat{y}(p_i) \in \{1, \dots, K\} \) be the predicted cluster label for that pixel.

We define the pixel-wise overlap between each predicted cluster \( k \in \{1, \dots, K\} \) and each ground-truth class \( c \in \{1, \dots, C\} \) as:
\begin{equation}
    O_{k,c} = \left| \left\{ p_i \in \mathcal{P} \;\middle|\; \hat{y}(p_i) = k \wedge y(p_i) = c \right\} \right|
\end{equation}

We then define the cluster-to-class mapping \( \pi: \{1, \dots, K\} \rightarrow \{1, \dots, C\} \) using majority voting:
\begin{equation}
    \pi(k) = \argmax_{c \in \{1, \dots, C\}} O_{k,c}
\end{equation}

This many-to-one mapping allows multiple predicted clusters to be associated with the same ground-truth class, accommodating over-segmentation.

Unlike methods based on the Hungarian algorithm, which solve a one-to-one assignment problem using a cost matrix (often derived from IoU), our overlap-based majority voting approach directly captures the dominant semantic association for each cluster. This enables flexible evaluation even when the output is fragmented or highly redundant.

In the \textbf{temporal case}, the mapping \( \pi \) is computed once over the entire video sequence and then held fixed for all frames during metric computation, ensuring temporal consistency.

In the \textbf{zero-shot case}, the mapping \( \pi_t \) is computed independently for each frame \( t \), based only on the overlaps observed in that frame. This permits flexible frame-level evaluation without assuming any temporal structure.

\subsection{Metrics}
\paragraph{Mean Intersection over Union (mIoU):}
\begin{equation}
    \text{mIoU} = \frac{1}{C} \sum_{c=1}^{C} \frac{TP_c}{TP_c + FP_c + FN_c}
\end{equation}
Where:
\begin{itemize}
    \item $C$ is the number of classes (after Hungarian matching)
    \item $TP_c$ is the number of true positive pixels for class $c$
    \item $FP_c$ and $FN_c$ are the numbers of false positive and false negative pixels for class $c$, respectively
\end{itemize}


\paragraph{Mean Pixel Accuracy (Acc):}
\begin{equation}
    \text{Acc} = \frac{1}{C} \sum_{c=1}^{C} \frac{TP_c}{TP_c + FN_c}
\end{equation}
Where:
\begin{itemize}
    \item $TP_c$ denotes the number of pixels correctly predicted as class $c$
    \item $FN_c$ denotes the number of ground truth pixels of class $c$ that were not correctly predicted
\end{itemize}


\paragraph{Note on Class Imbalance:}
In datasets with class imbalance—where some classes dominate the pixel distribution—overall metrics such as IoU and pixel accuracy can be misleading, as they are heavily influenced by large classes.
In contrast, mean metrics such as mIoU and mAcc treat each class equally, providing a more balanced evaluation of segmentation performance across all classes.
Therefore, we primarily report mIoU and mAcc to better reflect performance on rare or underrepresented classes.
This evaluation strategy is consistent with standard benchmarks such as PASCAL VOC~\cite{everingham2010pascal} and Cityscapes~\cite{cordts2016cityscapes}, which report class-averaged scores as the primary metric.

\paragraph{Over- and Under-Segmentation Entropy}
While mIoU and pixel accuracy are standard metrics, they can obscure important structural properties of a segmentation.
For instance, an algorithm may receive a low mIoU despite correctly identifying small structures (due to over-fragmentation), or a deceptively high score by over-smoothing regions (under-segmentation).
To better characterize such behavior, we measure the \textit{over-segmentation entropy} and \textit{under-segmentation entropy}, which quantify the uncertainty in one labeling conditioned on the other.

Let \( \mathcal{P} = \{p_1, \dots, p_N\} \) be the set of all pixels in the dataset. Each pixel has a ground-truth class label \( y(p_i) \in \{1, \dots, C\} \) and a predicted cluster label \( \hat{y}(p_i) \in \{1, \dots, K\} \).

We define the marginal distribution over predicted clusters as:
\begin{equation}
    P(\hat{y} = k) = \frac{|\{ p_i \in \mathcal{P} \mid \hat{y}(p_i) = k \}|}{|\mathcal{P}|}
\end{equation}

The joint distribution between predicted clusters and ground-truth classes is:
\begin{equation}
    P(\hat{y} = k, \; y = c) = \frac{|\{ p_i \in \mathcal{P} \mid \hat{y}(p_i) = k \wedge y(p_i) = c \}|}{|\mathcal{P}|}
\end{equation}

From this, the conditional distributions are:
\begin{equation}
    P(\hat{y} = k \mid y = c) = \frac{P(\hat{y} = k, \; y = c)}{P(y = c)}, \qquad
    P(y = c \mid \hat{y} = k) = \frac{P(\hat{y} = k, \; y = c)}{P(\hat{y} = k)}
\end{equation}

The over-segmentation entropy, which captures how fragmented each semantic class is across clusters, is given by:
\begin{equation}
    \mathcal{H}(\hat{y} \mid y) = -\sum_{c=1}^{C} \sum_{k=1}^{K} P(y = c, \hat{y} = k) \log P(\hat{y} = k \mid y = c)
\end{equation}

Similarly, the under-segmentation entropy, which captures how many semantic classes are merged into each predicted cluster, is given by:
\begin{equation}
    \mathcal{H}(y \mid \hat{y}) = -\sum_{k=1}^{K} \sum_{c=1}^{C} P(\hat{y} = k, y = c) \log P(y = c \mid \hat{y} = k)
\end{equation}

Both entropy values lie in the range \([0, \log K]\) or \([0, \log C]\), and are typically normalized to \([0, 1]\) for interpretability. 
Lower values indicate more faithful correspondence between predictions and ground truth, while higher values suggest either excessive fragmentation (\(\mathcal{H}(\hat{y} \mid y)\)) or semantic collapse (\(\mathcal{H}(y \mid \hat{y})\)). 
In practice, high over-segmentation entropy reflects that individual semantic classes are being split across many predicted clusters, complicating downstream reasoning, while high under-segmentation entropy indicates that predicted clusters conflate multiple semantic classes, reducing discriminative power.

\newpage

\subsection{RUGD - Temporal}
\setlength{\tabcolsep}{3pt}
\begin{table}[h]
\centering
\caption{mIoU and Acc Scores (\%) by model, then by subdataset for each number of clusters (all models)}\vspace{0.5em}
\label{tab:miou_pixelacc_all_models_RUGD_TEMPORAL_1}
\begin{tabular}{lrrrrrrrrrrrr}
\toprule
 & \multicolumn{2}{c}{\textbf{\boldmath$k=400$}} & \multicolumn{2}{c}{\textbf{\boldmath$k=200$}} & \multicolumn{2}{c}{\textbf{\boldmath$k=100$}} & \multicolumn{2}{c}{\textbf{\boldmath$k=50$}} & \multicolumn{2}{c}{\textbf{\boldmath$k=25$}} & \multicolumn{2}{c}{\textbf{\boldmath$k=12$}} \\

\cmidrule(lr){2-3}
\cmidrule(lr){4-5}
\cmidrule(lr){6-7}
\cmidrule(lr){8-9}
\cmidrule(lr){10-11}
\cmidrule(lr){12-13}
  & \multicolumn{1}{c}{\textbf{mIoU}} & \multicolumn{1}{c}{\textbf{Acc}} & \multicolumn{1}{c}{\textbf{mIoU}} & \multicolumn{1}{c}{\textbf{Acc}} & \multicolumn{1}{c}{\textbf{mIoU}} & \multicolumn{1}{c}{\textbf{Acc}} & \multicolumn{1}{c}{\textbf{mIoU}} & \multicolumn{1}{c}{\textbf{Acc}} & \multicolumn{1}{c}{\textbf{mIoU}} & \multicolumn{1}{c}{\textbf{Acc}} & \multicolumn{1}{c}{\textbf{mIoU}} & \multicolumn{1}{c}{\textbf{Acc}} \\[0.4ex]
\midrule
\textbf{Frontier} \\ \small{(DinoV2-B)} &  &  &  &  &  &  &  &  &  &  &  &  \\
creek & 52.90 & 84.50 & 46.80 & 81.50 & 44.70 & 79.50 & 39.30 & 76.20 & 28.30 & 69.60 & 23.50 & 67.50 \\
trail & 36.00 & 88.00 & 33.90 & 86.70 & 32.50 & 85.60 & 31.10 & 84.90 & 27.20 & 80.40 & 18.00 & 74.40 \\
village & 51.20 & 89.20 & 51.10 & 87.70 & 48.70 & 86.60 & 39.30 & 83.70 & 46.30 & 83.00 & 35.30 & 79.80 \\
park-1 & 50.20 & 86.30 & 46.50 & 84.50 & 44.30 & 82.60 & 38.70 & 78.70 & 33.00 & 77.60 & 20.70 & 68.80 \\
\bottomrule
\end{tabular}
\end{table}
\clearpage
\setlength{\tabcolsep}{3pt}
\begin{table}[h]
\centering
\caption{mIoU and Acc Scores (\%) by model, then by subdataset for each number of clusters (all models)}\vspace{0.5em}
\label{tab:miou_pixelacc_all_models_RUGD_TEMPORAL_2}
\begin{tabular}{lrrrrrrrrrrrr}
\toprule
 & \multicolumn{2}{c}{\textbf{\boldmath$k=400$}} & \multicolumn{2}{c}{\textbf{\boldmath$k=200$}} & \multicolumn{2}{c}{\textbf{\boldmath$k=100$}} & \multicolumn{2}{c}{\textbf{\boldmath$k=50$}} & \multicolumn{2}{c}{\textbf{\boldmath$k=25$}} & \multicolumn{2}{c}{\textbf{\boldmath$k=12$}} \\

\cmidrule(lr){2-3}
\cmidrule(lr){4-5}
\cmidrule(lr){6-7}
\cmidrule(lr){8-9}
\cmidrule(lr){10-11}
\cmidrule(lr){12-13}
  & \multicolumn{1}{c}{\textbf{mIoU}} & \multicolumn{1}{c}{\textbf{Acc}} & \multicolumn{1}{c}{\textbf{mIoU}} & \multicolumn{1}{c}{\textbf{Acc}} & \multicolumn{1}{c}{\textbf{mIoU}} & \multicolumn{1}{c}{\textbf{Acc}} & \multicolumn{1}{c}{\textbf{mIoU}} & \multicolumn{1}{c}{\textbf{Acc}} & \multicolumn{1}{c}{\textbf{mIoU}} & \multicolumn{1}{c}{\textbf{Acc}} & \multicolumn{1}{c}{\textbf{mIoU}} & \multicolumn{1}{c}{\textbf{Acc}} \\[0.4ex]
\midrule
\textbf{DiffCut} \\ \small(${\tau=0.90}$) &  &  &  &  &  &  &  &  &  &  &  &  \\
creek & 15.40 & 59.40 & 14.90 & 57.90 & 12.90 & 56.00 & 10.40 & 52.50 & 10.00 & 50.40 & 6.60 & 48.10 \\
trail & 13.60 & 65.20 & 10.60 & 63.40 & 9.70 & 61.50 & 8.30 & 58.70 & 5.80 & 56.70 & 5.20 & 56.20 \\
village & 17.10 & 64.10 & 14.40 & 61.40 & 13.80 & 59.50 & 13.50 & 59.30 & 13.70 & 58.60 & 8.10 & 54.00 \\
park-1 & 16.10 & 64.20 & 15.00 & 62.20 & 12.90 & 61.00 & 11.90 & 58.80 & 8.50 & 55.40 & 7.10 & 53.50 \\
\bottomrule
\end{tabular}
\end{table}
\clearpage
\setlength{\tabcolsep}{3pt}
\begin{table}[h]
\centering
\caption{OSE and USE Scores by Model, then by Subdataset for Each Number of Clusters (All Models)}\vspace{0.5em}
\label{tab:ose_use_all_models_RUGD_TEMPORAL1}
\begin{tabular}{lrrrrrrrrrrrr}
\toprule
 & \multicolumn{2}{c}{\textbf{\boldmath$k=400$}} & \multicolumn{2}{c}{\textbf{\boldmath$k=200$}} & \multicolumn{2}{c}{\textbf{\boldmath$k=100$}} & \multicolumn{2}{c}{\textbf{\boldmath$k=50$}} & \multicolumn{2}{c}{\textbf{\boldmath$k=25$}} & \multicolumn{2}{c}{\textbf{\boldmath$k=12$}} \\

\cmidrule(lr){2-3}
\cmidrule(lr){4-5}
\cmidrule(lr){6-7}
\cmidrule(lr){8-9}
\cmidrule(lr){10-11}
\cmidrule(lr){12-13}
  & \multicolumn{1}{c}{\textbf{OSE}} & \multicolumn{1}{c}{\textbf{USE}} & \multicolumn{1}{c}{\textbf{OSE}} & \multicolumn{1}{c}{\textbf{USE}} & \multicolumn{1}{c}{\textbf{OSE}} & \multicolumn{1}{c}{\textbf{USE}} & \multicolumn{1}{c}{\textbf{OSE}} & \multicolumn{1}{c}{\textbf{USE}} & \multicolumn{1}{c}{\textbf{OSE}} & \multicolumn{1}{c}{\textbf{USE}} & \multicolumn{1}{c}{\textbf{OSE}} & \multicolumn{1}{c}{\textbf{USE}} \\[0.4ex]
\midrule
\textbf{Frontier} \\ \small{(DinoV2-B)} &  &  &  &  &  &  &  &  &  &  &  &  \\
creek & 1.41 & 0.33 & 1.13 & 0.39 & 0.88 & 0.45 & 0.63 & 0.51 & 0.43 & 0.58 & 0.31 & 0.62 \\
trail & 1.87 & 0.23 & 1.59 & 0.25 & 1.35 & 0.28 & 1.08 & 0.31 & 0.87 & 0.38 & 0.53 & 0.45 \\
village & 1.07 & 0.27 & 0.99 & 0.29 & 0.88 & 0.30 & 0.77 & 0.32 & 0.64 & 0.38 & 0.57 & 0.39 \\
park-1 & 1.52 & 0.30 & 1.16 & 0.35 & 0.86 & 0.39 & 0.68 & 0.44 & 0.57 & 0.46 & 0.44 & 0.59 \\
\bottomrule
\end{tabular}
\end{table}
\clearpage
\setlength{\tabcolsep}{3pt}
\begin{table}[h]
\centering
\caption{OSE and USE Scores by Model, then by Subdataset for Each Number of Clusters (All Models)}\vspace{0.5em}
\label{tab:ose_use_all_models_RUGD_TEMPORAL2}
\begin{tabular}{lrrrrrrrrrrrr}
\toprule
 & \multicolumn{2}{c}{\textbf{\boldmath$k=400$}} & \multicolumn{2}{c}{\textbf{\boldmath$k=200$}} & \multicolumn{2}{c}{\textbf{\boldmath$k=100$}} & \multicolumn{2}{c}{\textbf{\boldmath$k=50$}} & \multicolumn{2}{c}{\textbf{\boldmath$k=25$}} & \multicolumn{2}{c}{\textbf{\boldmath$k=12$}} \\

\cmidrule(lr){2-3}
\cmidrule(lr){4-5}
\cmidrule(lr){6-7}
\cmidrule(lr){8-9}
\cmidrule(lr){10-11}
\cmidrule(lr){12-13}
  & \multicolumn{1}{c}{\textbf{OSE}} & \multicolumn{1}{c}{\textbf{USE}} & \multicolumn{1}{c}{\textbf{OSE}} & \multicolumn{1}{c}{\textbf{USE}} & \multicolumn{1}{c}{\textbf{OSE}} & \multicolumn{1}{c}{\textbf{USE}} & \multicolumn{1}{c}{\textbf{OSE}} & \multicolumn{1}{c}{\textbf{USE}} & \multicolumn{1}{c}{\textbf{OSE}} & \multicolumn{1}{c}{\textbf{USE}} & \multicolumn{1}{c}{\textbf{OSE}} & \multicolumn{1}{c}{\textbf{USE}} \\[0.4ex]
\midrule
\textbf{DiffCut} \\ \small(${\tau=0.90}$) &  &  &  &  &  &  &  &  &  &  &  &  \\
creek & 1.02 & 0.91 & 0.84 & 0.93 & 0.70 & 0.96 & 0.61 & 0.98 & 0.51 & 1.01 & 0.42 & 1.03 \\
trail & 1.19 & 0.72 & 0.99 & 0.75 & 0.86 & 0.77 & 0.76 & 0.80 & 0.68 & 0.82 & 0.60 & 0.83 \\
village & 1.06 & 0.90 & 0.93 & 0.93 & 0.87 & 0.94 & 0.80 & 0.96 & 0.77 & 0.97 & 0.66 & 1.00 \\
park-1 & 1.22 & 0.86 & 1.04 & 0.91 & 0.91 & 0.94 & 0.78 & 0.97 & 0.70 & 0.99 & 0.60 & 1.03 \\
\bottomrule
\end{tabular}
\end{table}
\clearpage

\subsection{RUGD - Zero Shot}
\setlength{\tabcolsep}{3pt}
\begin{table}[h]
\centering
\caption{mIoU and Acc Scores (\%) by model, then by subdataset for each number of clusters (all models)}\vspace{0.5em}
\label{tab:miou_pixelacc_all_models_RUGD_ZEROSHOT}
\begin{tabular}{lrrrrrrrrrrrr}
\toprule
 & \multicolumn{2}{c}{\textbf{\boldmath$k=400$}} & \multicolumn{2}{c}{\textbf{\boldmath$k=200$}} & \multicolumn{2}{c}{\textbf{\boldmath$k=100$}} & \multicolumn{2}{c}{\textbf{\boldmath$k=50$}} & \multicolumn{2}{c}{\textbf{\boldmath$k=25$}} & \multicolumn{2}{c}{\textbf{\boldmath$k=12$}} \\

\cmidrule(lr){2-3}
\cmidrule(lr){4-5}
\cmidrule(lr){6-7}
\cmidrule(lr){8-9}
\cmidrule(lr){10-11}
\cmidrule(lr){12-13}
  & \multicolumn{1}{c}{\textbf{mIoU}} & \multicolumn{1}{c}{\textbf{Acc}} & \multicolumn{1}{c}{\textbf{mIoU}} & \multicolumn{1}{c}{\textbf{Acc}} & \multicolumn{1}{c}{\textbf{mIoU}} & \multicolumn{1}{c}{\textbf{Acc}} & \multicolumn{1}{c}{\textbf{mIoU}} & \multicolumn{1}{c}{\textbf{Acc}} & \multicolumn{1}{c}{\textbf{mIoU}} & \multicolumn{1}{c}{\textbf{Acc}} & \multicolumn{1}{c}{\textbf{mIoU}} & \multicolumn{1}{c}{\textbf{Acc}} \\[0.4ex]
\midrule
\textbf{Frontier} \\ \small{(DinoV2-B)} &  &  &  &  &  &  &  &  &  &  &  &  \\
creek & 53.20 & 88.30 & 52.70 & 88.10 & 52.50 & 88.00 & 52.20 & 87.90 & 52.20 & 87.80 & 52.10 & 87.80 \\
trail & 56.00 & 91.50 & 55.70 & 91.40 & 55.60 & 91.40 & 55.50 & 91.30 & 55.40 & 91.30 & 55.40 & 91.30 \\
village & 71.80 & 94.60 & 71.50 & 94.50 & 71.30 & 94.50 & 71.20 & 94.40 & 71.20 & 94.30 & 71.20 & 94.30 \\
park-1 & 58.90 & 90.70 & 58.50 & 90.60 & 58.20 & 90.50 & 57.70 & 90.50 & 57.50 & 90.40 & 57.20 & 90.40 \\
\bottomrule
\end{tabular}
\end{table}
\clearpage
\setlength{\tabcolsep}{3pt}
\begin{table}[h]
\centering
\caption{mIoU and Acc Scores (\%) by Model, then by Subdataset (DiffCut Models - Zero Shot)}\vspace{0.5em}
\label{tab:miou_pixelacc_simplified_RUGD_ZEROSHOT}
\begin{tabular}{lrr}
\toprule
 & \multicolumn{1}{c}{\textbf{mIoU}} & \multicolumn{1}{c}{\textbf{Acc}} \\[0.4ex]
\midrule
\textbf{DiffCut} \\ \small(${\tau=0.90}$) &  &  \\
creek & 53.20 & 91.10 \\
trail & 57.20 & 93.00 \\
village & 50.50 & 93.00 \\
park-1 & 44.60 & 89.90 \\
park-2 & 46.60 & 89.60 \\
park-8 & 41.50 & 90.10 \\
trail-3 & 53.80 & 92.30 \\
trail-4 & 49.70 & 90.40 \\
trail-5 & 53.20 & 91.00 \\
trail-6 & 49.50 & 89.20 \\
trail-7 & 48.10 & 90.40 \\
trail-9 & 49.00 & 91.40 \\
trail-10 & 58.60 & 89.30 \\
trail-11 & 53.00 & 89.80 \\
trail-12 & 52.30 & 89.90 \\
trail-13 & 49.70 & 88.90 \\
trail-14 & 54.20 & 90.40 \\
trail-15 & 53.40 & 87.70 \\
\textbf{Average} & \textbf{51.01} & \textbf{90.41} \\
\midrule &  &  \\
\textbf{DiffCut} \\ \small(${\tau=0.95}$) &  &  \\
creek & 55.40 & 92.20 \\
trail & 59.70 & 93.90 \\
village & 51.70 & 93.60 \\
park-1 & 46.40 & 91.00 \\
park-2 & 48.90 & 90.90 \\
park-8 & 43.60 & 91.10 \\
trail-3 & 56.20 & 93.20 \\
trail-4 & 53.20 & 91.70 \\
trail-5 & 55.60 & 91.90 \\
trail-6 & 52.70 & 90.50 \\
trail-7 & 50.30 & 91.10 \\
trail-9 & 52.60 & 92.70 \\
trail-10 & 64.10 & 91.10 \\
trail-11 & 56.30 & 91.40 \\
trail-12 & 55.70 & 91.50 \\
trail-13 & 53.10 & 91.10 \\
trail-14 & 57.70 & 91.90 \\
trail-15 & 57.50 & 89.90 \\
\textbf{Average} & \textbf{53.93} & \textbf{91.71} \\
\bottomrule
\end{tabular}
\end{table}
\clearpage
\setlength{\tabcolsep}{3pt}
\begin{table}[h]
\centering
\caption{OSE and USE Scores by Model, then by Subdataset for Each Number of Clusters (All Models)}\vspace{0.5em}
\label{tab:ose_use_all_models_RUGD_ZEROSHOT}
\begin{tabular}{lrrrrrrrrrrrr}
\toprule
 & \multicolumn{2}{c}{\textbf{\boldmath$k=400$}} & \multicolumn{2}{c}{\textbf{\boldmath$k=200$}} & \multicolumn{2}{c}{\textbf{\boldmath$k=100$}} & \multicolumn{2}{c}{\textbf{\boldmath$k=50$}} & \multicolumn{2}{c}{\textbf{\boldmath$k=25$}} & \multicolumn{2}{c}{\textbf{\boldmath$k=12$}} \\

\cmidrule(lr){2-3}
\cmidrule(lr){4-5}
\cmidrule(lr){6-7}
\cmidrule(lr){8-9}
\cmidrule(lr){10-11}
\cmidrule(lr){12-13}
  & \multicolumn{1}{c}{\textbf{OSE}} & \multicolumn{1}{c}{\textbf{USE}} & \multicolumn{1}{c}{\textbf{OSE}} & \multicolumn{1}{c}{\textbf{USE}} & \multicolumn{1}{c}{\textbf{OSE}} & \multicolumn{1}{c}{\textbf{USE}} & \multicolumn{1}{c}{\textbf{OSE}} & \multicolumn{1}{c}{\textbf{USE}} & \multicolumn{1}{c}{\textbf{OSE}} & \multicolumn{1}{c}{\textbf{USE}} & \multicolumn{1}{c}{\textbf{OSE}} & \multicolumn{1}{c}{\textbf{USE}} \\[0.4ex]
\midrule
\textbf{Frontier} \\ \small{(DinoV2-B)} &  &  &  &  &  &  &  &  &  &  &  &  \\
creek & 2.28 & 0.29 & 2.20 & 0.30 & 2.15 & 0.30 & 2.12 & 0.30 & 2.09 & 0.31 & 2.07 & 0.31 \\
trail & 2.66 & 0.21 & 2.60 & 0.22 & 2.57 & 0.22 & 2.54 & 0.22 & 2.52 & 0.22 & 2.50 & 0.22 \\
village & 2.31 & 0.14 & 2.27 & 0.15 & 2.25 & 0.15 & 2.22 & 0.15 & 2.20 & 0.15 & 2.18 & 0.15 \\
park-1 & 2.37 & 0.24 & 2.30 & 0.24 & 2.27 & 0.25 & 2.25 & 0.25 & 2.23 & 0.25 & 2.21 & 0.25 \\
\bottomrule
\end{tabular}
\end{table}
\clearpage
\setlength{\tabcolsep}{3pt}
\begin{table}[h]
\centering
\caption{OSE and USE Scores by Model, then by Subdataset (DiffCut Models - Zero Shot)}\vspace{0.5em}
\label{tab:ose_use_simplified_RUGD_ZEROSHOT}
\begin{tabular}{lrr}
\toprule
 & \multicolumn{1}{c}{\textbf{OSE}} & \multicolumn{1}{c}{\textbf{USE}} \\[0.4ex]
\midrule
\textbf{DiffCut} \\ \small(${\tau=0.90}$) &  &  \\
creek & 2.96 & 0.21 \\
trail & 2.87 & 0.17 \\
village & 2.34 & 0.16 \\
park-1 & 2.41 & 0.24 \\
park-2 & 2.29 & 0.25 \\
park-8 & 2.48 & 0.24 \\
trail-3 & 2.68 & 0.18 \\
trail-4 & 2.58 & 0.23 \\
trail-5 & 2.74 & 0.22 \\
trail-6 & 2.60 & 0.26 \\
trail-7 & 2.71 & 0.23 \\
trail-9 & 2.16 & 0.21 \\
trail-10 & 2.20 & 0.26 \\
trail-11 & 2.33 & 0.24 \\
trail-12 & 2.44 & 0.24 \\
trail-13 & 2.29 & 0.26 \\
trail-14 & 2.51 & 0.23 \\
trail-15 & 2.38 & 0.28 \\
\textbf{Average} & \textbf{2.50} & \textbf{0.23} \\
\midrule &  &  \\
\textbf{DiffCut} \\ \small(${\tau=0.95}$) &  &  \\
creek & 3.88 & 0.18 \\
trail & 3.97 & 0.14 \\
village & 2.98 & 0.15 \\
park-1 & 3.32 & 0.21 \\
park-2 & 3.20 & 0.21 \\
park-8 & 3.49 & 0.21 \\
trail-3 & 3.74 & 0.15 \\
trail-4 & 3.70 & 0.19 \\
trail-5 & 3.82 & 0.19 \\
trail-6 & 3.74 & 0.22 \\
trail-7 & 3.84 & 0.20 \\
trail-9 & 3.30 & 0.17 \\
trail-10 & 3.38 & 0.21 \\
trail-11 & 3.40 & 0.20 \\
trail-12 & 3.54 & 0.20 \\
trail-13 & 3.31 & 0.20 \\
trail-14 & 3.64 & 0.19 \\
trail-15 & 3.45 & 0.23 \\
\textbf{Average} & \textbf{3.54} & \textbf{0.19} \\
\bottomrule
\end{tabular}
\end{table}
\clearpage

\subsection{RELLIS - Temporal}
\setlength{\tabcolsep}{3pt}
\begin{table}[h]
\centering
\caption{mIoU and Acc Scores (\%) by model, then by subdataset for each number of clusters (all models)}\vspace{0.5em}
\label{tab:miou_pixelacc_all_models_RELLIS_TEMPORAL}
\begin{tabular}{lrrrrrrrrrrrr}
\toprule
 & \multicolumn{2}{c}{\textbf{\boldmath$k=400$}} & \multicolumn{2}{c}{\textbf{\boldmath$k=200$}} & \multicolumn{2}{c}{\textbf{\boldmath$k=100$}} & \multicolumn{2}{c}{\textbf{\boldmath$k=50$}} & \multicolumn{2}{c}{\textbf{\boldmath$k=25$}} & \multicolumn{2}{c}{\textbf{\boldmath$k=12$}} \\

\cmidrule(lr){2-3}
\cmidrule(lr){4-5}
\cmidrule(lr){6-7}
\cmidrule(lr){8-9}
\cmidrule(lr){10-11}
\cmidrule(lr){12-13}
  & \multicolumn{1}{c}{\textbf{mIoU}} & \multicolumn{1}{c}{\textbf{Acc}} & \multicolumn{1}{c}{\textbf{mIoU}} & \multicolumn{1}{c}{\textbf{Acc}} & \multicolumn{1}{c}{\textbf{mIoU}} & \multicolumn{1}{c}{\textbf{Acc}} & \multicolumn{1}{c}{\textbf{mIoU}} & \multicolumn{1}{c}{\textbf{Acc}} & \multicolumn{1}{c}{\textbf{mIoU}} & \multicolumn{1}{c}{\textbf{Acc}} & \multicolumn{1}{c}{\textbf{mIoU}} & \multicolumn{1}{c}{\textbf{Acc}} \\[0.4ex]
\midrule
\textbf{Frontier} \\ \small{(DinoV2-B)} &  &  &  &  &  &  &  &  &  &  &  &  \\
00000 & 31.80 & 80.50 & 30.40 & 79.50 & 29.20 & 78.20 & 24.40 & 75.50 & 23.00 & 74.00 & 21.00 & 71.50 \\
00001 & 37.30 & 86.10 & 36.00 & 84.90 & 34.40 & 83.90 & 34.50 & 83.60 & 33.30 & 82.40 & 30.80 & 80.40 \\
00002 & 25.40 & 88.40 & 26.60 & 87.10 & 22.60 & 86.30 & 21.30 & 84.20 & 20.20 & 82.40 & 13.60 & 68.20 \\
00003 & 36.30 & 85.30 & 35.00 & 84.50 & 33.00 & 82.90 & 32.40 & 81.20 & 29.60 & 80.10 & 28.80 & 78.80 \\
\bottomrule
\end{tabular}
\end{table}
\clearpage
\setlength{\tabcolsep}{3pt}
\begin{table}[h]
\centering
\caption{OSE and USE Scores by Model, then by Subdataset for Each Number of Clusters (All Models)}\vspace{0.5em}
\label{tab:ose_use_all_models_RELLIS_TEMPORAL}
\begin{tabular}{lrrrrrrrrrrrr}
\toprule
 & \multicolumn{2}{c}{\textbf{\boldmath$k=400$}} & \multicolumn{2}{c}{\textbf{\boldmath$k=200$}} & \multicolumn{2}{c}{\textbf{\boldmath$k=100$}} & \multicolumn{2}{c}{\textbf{\boldmath$k=50$}} & \multicolumn{2}{c}{\textbf{\boldmath$k=25$}} & \multicolumn{2}{c}{\textbf{\boldmath$k=12$}} \\

\cmidrule(lr){2-3}
\cmidrule(lr){4-5}
\cmidrule(lr){6-7}
\cmidrule(lr){8-9}
\cmidrule(lr){10-11}
\cmidrule(lr){12-13}
  & \multicolumn{1}{c}{\textbf{OSE}} & \multicolumn{1}{c}{\textbf{USE}} & \multicolumn{1}{c}{\textbf{OSE}} & \multicolumn{1}{c}{\textbf{USE}} & \multicolumn{1}{c}{\textbf{OSE}} & \multicolumn{1}{c}{\textbf{USE}} & \multicolumn{1}{c}{\textbf{OSE}} & \multicolumn{1}{c}{\textbf{USE}} & \multicolumn{1}{c}{\textbf{OSE}} & \multicolumn{1}{c}{\textbf{USE}} & \multicolumn{1}{c}{\textbf{OSE}} & \multicolumn{1}{c}{\textbf{USE}} \\[0.4ex]
\midrule
\textbf{Frontier} \\ \small{(DinoV2-B)} &  &  &  &  &  &  &  &  &  &  &  &  \\
00000 & 2.17 & 0.42 & 1.87 & 0.45 & 1.58 & 0.48 & 1.29 & 0.54 & 1.07 & 0.58 & 0.81 & 0.64 \\
00001 & 2.13 & 0.33 & 1.80 & 0.36 & 1.45 & 0.39 & 1.22 & 0.41 & 0.95 & 0.44 & 0.57 & 0.51 \\
00002 & 1.93 & 0.26 & 1.55 & 0.29 & 1.22 & 0.33 & 0.89 & 0.39 & 0.66 & 0.42 & 0.43 & 0.57 \\
00003 & 2.34 & 0.34 & 2.03 & 0.37 & 1.74 & 0.39 & 1.49 & 0.43 & 1.23 & 0.47 & 0.95 & 0.51 \\
\bottomrule
\end{tabular}
\end{table}
\clearpage

\subsection{RELLIS - Zero-Shot}
\setlength{\tabcolsep}{3pt}
\begin{table}[h]
\centering
\caption{mIoU and Acc Scores (\%) by model, then by subdataset for each number of clusters (all models)}\vspace{0.5em}
\label{tab:miou_pixelacc_all_models_RELLIS_ZEROSHOT}
\begin{tabular}{lrrrrrrrrrrrr}
\toprule
 & \multicolumn{2}{c}{\textbf{\boldmath$k=400$}} & \multicolumn{2}{c}{\textbf{\boldmath$k=200$}} & \multicolumn{2}{c}{\textbf{\boldmath$k=100$}} & \multicolumn{2}{c}{\textbf{\boldmath$k=50$}} & \multicolumn{2}{c}{\textbf{\boldmath$k=25$}} & \multicolumn{2}{c}{\textbf{\boldmath$k=12$}} \\

\cmidrule(lr){2-3}
\cmidrule(lr){4-5}
\cmidrule(lr){6-7}
\cmidrule(lr){8-9}
\cmidrule(lr){10-11}
\cmidrule(lr){12-13}
  & \multicolumn{1}{c}{\textbf{mIoU}} & \multicolumn{1}{c}{\textbf{Acc}} & \multicolumn{1}{c}{\textbf{mIoU}} & \multicolumn{1}{c}{\textbf{Acc}} & \multicolumn{1}{c}{\textbf{mIoU}} & \multicolumn{1}{c}{\textbf{Acc}} & \multicolumn{1}{c}{\textbf{mIoU}} & \multicolumn{1}{c}{\textbf{Acc}} & \multicolumn{1}{c}{\textbf{mIoU}} & \multicolumn{1}{c}{\textbf{Acc}} & \multicolumn{1}{c}{\textbf{mIoU}} & \multicolumn{1}{c}{\textbf{Acc}} \\[0.4ex]
\midrule
\textbf{Frontier} \\ \small{(DinoV2-B)} &  &  &  &  &  &  &  &  &  &  &  &  \\
00000 & 41.10 & 83.10 & 39.80 & 82.50 & 37.30 & 81.10 & 34.90 & 79.60 & 33.20 & 78.00 & 28.70 & 75.50 \\
00001 & 47.50 & 86.50 & 45.30 & 85.40 & 42.70 & 84.10 & 41.00 & 83.00 & 39.30 & 81.70 & 38.00 & 80.80 \\
00002 & 41.40 & 89.40 & 39.60 & 88.20 & 38.20 & 86.70 & 34.40 & 83.80 & 30.00 & 80.80 & 28.50 & 79.60 \\
00003 & 44.20 & 85.70 & 42.20 & 84.80 & 40.90 & 84.20 & 39.10 & 83.00 & 37.10 & 81.30 & 36.50 & 80.90 \\
\bottomrule
\end{tabular}
\end{table}
\setlength{\tabcolsep}{3pt}
\begin{table}[h]
\centering
\caption{mIoU and Acc Scores (\%) by Model, then by Subdataset (DiffCut Models - Zero Shot)}\vspace{0.5em}
\label{tab:miou_pixelacc_simplified_RELLIS_ZEROSHOT}
\begin{tabular}{lrr}
\toprule
 & \multicolumn{1}{c}{\textbf{mIoU}} & \multicolumn{1}{c}{\textbf{Acc}} \\[0.4ex]
\midrule
\textbf{DiffCut} \\ \small(${\tau=0.90}$) &  &  \\
00000 & 42.80 & 84.60 \\
00001 & 48.40 & 86.80 \\
00002 & 43.20 & 89.00 \\
00003 & 47.20 & 87.20 \\
00004 & 49.40 & 84.00 \\
\textbf{Average} & \textbf{46.20} & \textbf{86.32} \\
\midrule &  &  \\
\textbf{DiffCut} \\ \small(${\tau=0.95}$) &  &  \\
00000 & 46.80 & 86.60 \\
00001 & 51.50 & 88.20 \\
00002 & 45.50 & 90.80 \\
00003 & 50.00 & 88.40 \\
00004 & 53.10 & 86.00 \\
\textbf{Average} & \textbf{49.38} & \textbf{88.00} \\
\bottomrule
\end{tabular}
\end{table}
\clearpage
\setlength{\tabcolsep}{3pt}
\begin{table}[h]
\centering
\caption{OSE and USE Scores by Model, then by Subdataset for Each Number of Clusters (All Models)}\vspace{0.5em}
\label{tab:ose_use_all_models_RELLIS_ZEROSHOT}
\begin{tabular}{lrrrrrrrrrrrr}
\toprule
 & \multicolumn{2}{c}{\textbf{\boldmath$k=400$}} & \multicolumn{2}{c}{\textbf{\boldmath$k=200$}} & \multicolumn{2}{c}{\textbf{\boldmath$k=100$}} & \multicolumn{2}{c}{\textbf{\boldmath$k=50$}} & \multicolumn{2}{c}{\textbf{\boldmath$k=25$}} & \multicolumn{2}{c}{\textbf{\boldmath$k=12$}} \\

\cmidrule(lr){2-3}
\cmidrule(lr){4-5}
\cmidrule(lr){6-7}
\cmidrule(lr){8-9}
\cmidrule(lr){10-11}
\cmidrule(lr){12-13}
  & \multicolumn{1}{c}{\textbf{OSE}} & \multicolumn{1}{c}{\textbf{USE}} & \multicolumn{1}{c}{\textbf{OSE}} & \multicolumn{1}{c}{\textbf{USE}} & \multicolumn{1}{c}{\textbf{OSE}} & \multicolumn{1}{c}{\textbf{USE}} & \multicolumn{1}{c}{\textbf{OSE}} & \multicolumn{1}{c}{\textbf{USE}} & \multicolumn{1}{c}{\textbf{OSE}} & \multicolumn{1}{c}{\textbf{USE}} & \multicolumn{1}{c}{\textbf{OSE}} & \multicolumn{1}{c}{\textbf{USE}} \\[0.4ex]
\midrule
\textbf{Frontier} \\ \small{(DinoV2-B)} &  &  &  &  &  &  &  &  &  &  &  &  \\
00000 & 2.13 & 0.45 & 1.82 & 0.48 & 1.54 & 0.52 & 1.28 & 0.56 & 1.06 & 0.61 & 0.87 & 0.67 \\
00001 & 2.10 & 0.36 & 1.76 & 0.40 & 1.43 & 0.44 & 1.17 & 0.48 & 0.97 & 0.52 & 0.70 & 0.56 \\
00002 & 1.93 & 0.28 & 1.53 & 0.32 & 1.20 & 0.36 & 0.92 & 0.42 & 0.67 & 0.48 & 0.52 & 0.52 \\
00003 & 2.28 & 0.37 & 2.00 & 0.40 & 1.72 & 0.43 & 1.48 & 0.47 & 1.24 & 0.53 & 0.81 & 0.58 \\
\bottomrule
\end{tabular}
\end{table}
\setlength{\tabcolsep}{3pt}
\begin{table}[h]
\centering
\caption{OSE and USE Scores by Model, then by Subdataset (DiffCut Models - Zero Shot)}\vspace{0.5em}
\label{tab:ose_use_simplified_RELLIS_ZEROSHOT}
\begin{tabular}{lrr}
\toprule
 & \multicolumn{1}{c}{\textbf{OSE}} & \multicolumn{1}{c}{\textbf{USE}} \\[0.4ex]
\midrule
\textbf{DiffCut} \\ \small(${\tau=0.90}$) &  &  \\
00000 & 2.14 & 0.40 \\
00001 & 1.86 & 0.34 \\
00002 & 1.76 & 0.27 \\
00003 & 1.77 & 0.34 \\
00004 & 1.81 & 0.41 \\
\textbf{Average} & \textbf{1.87} & \textbf{0.35} \\
\midrule &  &  \\
\textbf{DiffCut} \\ \small(${\tau=0.95}$) &  &  \\
00000 & 3.18 & 0.33 \\
00001 & 2.81 & 0.29 \\
00002 & 2.66 & 0.22 \\
00003 & 2.67 & 0.29 \\
00004 & 2.74 & 0.35 \\
\textbf{Average} & \textbf{2.81} & \textbf{0.30} \\
\bottomrule
\end{tabular}
\end{table}
\clearpage







\end{document}